\documentclass[11pt,a4paper]{article}
\usepackage[hyperref]{acl2018}
\usepackage{times}
\usepackage{latexsym}
\usepackage{amsmath,amssymb}
\usepackage[compact]{titlesec}         
\usepackage{lipsum}

\usepackage{url}
\usepackage{graphicx}
\usepackage{multirow}
\usepackage{setspace}

\aclfinalcopy 

\title{Towards one-shot learning for rare-word translation with external experts}	

\author{Ngoc-Quan Pham \and Jan Niehues \and  Alex Waibel \\
		Karlsruhe Institute of Technology \\
        {\tt ngoc.pham@kit.edu jan.niehues@kit.edu alex.waibel@kit.edu} }

\date{}

\begin{document}
\maketitle
\begin{abstract}

Neural machine translation (NMT) has significantly improved the quality of automatic translation models. One of the main challenges in current systems is the translation of rare words. We present a generic approach to address this weakness by having external models annotate the training data as~\textbf{Experts}, and control the model-expert interaction with a pointer network and reinforcement learning. Our experiments using phrase-based models to simulate Experts to complement neural machine translation models show that the model can be trained to copy the annotations into the output consistently. We demonstrate the benefit of our proposed framework in out-of-domain translation scenarios with only lexical resources, improving more than 1.0 BLEU point in both translation directions English$\rightarrow$Spanish and German$\rightarrow$English. 

\end{abstract}

\section{Introduction}

Sequence to sequence models have recently become the state-of-the-art approach for machine translation~\cite{luong2015effective,vaswani2017attention}. This model architecture can directly approximate the conditional probability of the target sequence given a source sequence using neural networks~\cite{kalchbrenner2013recurrent}. As a result, not only do they model a smoother probability distribution~\cite{bengio2003neural} than the sparse phrase tables in statistical machine translation~\cite{koehn2003statistical}, but they can also jointly learn translation models, language models and even alignments in a single model~\cite{Bahdanau2014}.

One of the main weaknesses of neural machine translation models is poor handling of low frequency events. 
Neural models tend to prioritize output fluency over translation adequacy, and faced with rare words either silently ignore input~\cite{koehn2017six} or fall into under- or over-translation~\cite{tu2016modeling}. 
Examples of these situations include named entities, dates, and rare morphological forms. 
Improper handling of rare events can be harmful to industrial systems~\cite{wu2016google}, where translation mistakes can have serious ramifications. 
Similarly, translating in specific domains such as information technology or biology, a slight change in vocabulary can drastically alter meaning. It is important, then, to address translation of rare words.

While domain-specific parallel corpora can be used to adapt translation models efficiently~\cite{luong2015stanford}, parallel corpora for many domains can be difficult to collect, and this requires continued training. Translation lexicons, however, are much more commonly available. 
In this work, we introduce a strategy to incorporate external lexical knowledge, dubbed ``Expert annotation," into neural machine translation models. First, we annotate the lexical translations directly into the source side of the parallel data, so that the information is exposed during both training and inference. Second, inspired by CopyNet~\cite{gu2016incorporating}, we utilize a pointer network~\cite{vinyals2015pointer} to introduce a copy distribution over the source sentence, to increase the generation probability of rare words. Given that the expert annotation can differ from the reference, in order to encourage the model to copy the annotation we use reinforcement learning to guide the search, giving rewards when the annotation is used. Our work is motivated to be able to achieve One-Shot learning, which can help the model to accurately translate the events that are annotated during inference. Such ability can be transferred from an Expert which is capable of learning to translate lexically with one or few examples, such as dictionaries, or phrase-tables, or even human annotators.  

We realize our proposed framework with experiments on English$\rightarrow$Spanish and German$\rightarrow$English translation tasks. We focus on translation of rare events using translation suggestions from an Expert, here simulated by an additional phrase table. Specifically, we annotate rare words in our parallel data with best candidates from a phrase table before training, so that rare events are provided with suggested translations. Our model can be explicitly trained to copy the annotation approximately $90\%$ of the time, and it outperformed the baselines on translation accuracy of rare words, reaching up to $97\%$ accuracy. Also importantly, this performance is maintained when translating data in a different domain. Further analysis was done to verify the potential of our proposed framework.

\section{Background - Neural Machine Translation}
Neural machine translation (NMT) consists of an encoder and a decoder~\cite{Sutskever2014,vaswani2017attention} that directly approximate the conditional probability of a target sequence $Y = {y_1, y_2, \cdots, y_T}$ given a source sequence $X = {x_1, x_2, \cdots, x_M}$. The model is normally trained to maximize the log-likelihood of each target token given the previous words as well as the source sequence with respect to model parameters $\theta$ as in Equation~\ref{eq:nmt}:
\begin{equation}
\label{eq:nmt}
\begin{aligned}
\log P(Y|X; \theta) = \\ \Sigma_{t=1}^T (\log P(y_t | X, y_1, y_2, \cdots, y_t-1))
\end{aligned}
\end{equation}

\paragraph{The advantages of NMT} compared to phrased-based machine translation come from the neural architecture components:
\begin{itemize}
\item The embedding layers, which are shared between samples, allow the model to continuously represent discrete words and effectively capture word relationship~\cite{bengio2003neural,mikolov2013distributed}. Notably we refer to two different embedding layers being used in most models, one for the first input layer of the encoder/decoder, and another one at the decoder output layer that is used to compute the probability distribution (Equation~\ref{eq:nmt}).
\item Complex neural architectures like LSTMs~\cite{Hochreiter1997} or Transformers~\cite{vaswani2017attention} can represent structural sequences (sentences, phrases) effectively. 
\item Attention models~\cite{Bahdanau2014,luong2015effective} are capable of hierarchically modeling the translation mapping between sentence pairs. 
\end{itemize}

\paragraph{The challenges of NMT} These models are often attacked over their inability to learn to translate rare events, which are often named entities and rare morphological variants~\cite{arthur2016incorporating,koehn2017six,nguyen2017improving}. Learning from rare events is difficult due to the fact that the model parameters are not adequately updated. For example, the embeddings of the rare words are only updated a few times during training, and similarly for the patterns learned by the recurrent structures in the encoders / decoders and attention models.

\section{Expert framework description}
\label{sect:framework}
\begin{figure}
\caption{\label{fig:framework} A generic illustration of our framework. The source sentence is annotated with experts before learning. The model learns to utilize the annotation by using them directly in the translation)}
\includegraphics[width = 0.5\textwidth]{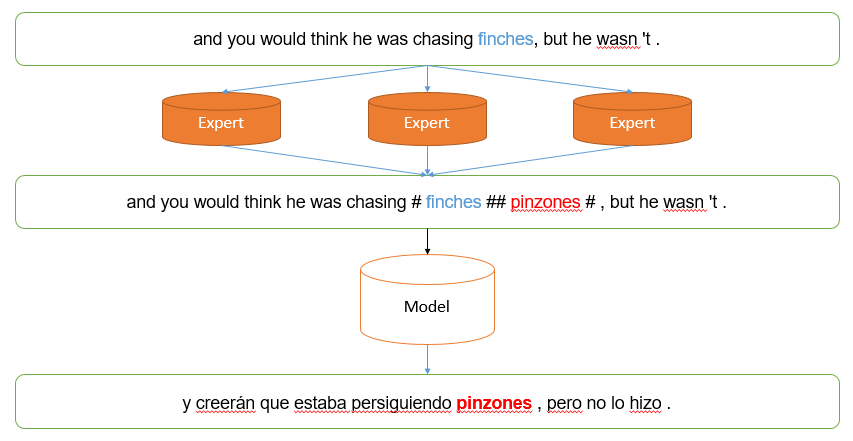}
\end{figure}

Human translators can benefit from external knowledge such as dictionaries, particularly in specific domains. Similarly, the idea behind our framework is to rely on external models to annotate extra input into the source side of the training data, which we refer as~\textbf{Experts}. Such expert models would not necessarily outperform NMT models themselves, but rather complement them and compensate for their weaknesses.

The illustration of the proposed framework is given in Figure~\ref{fig:framework}. Before the learning process, the source sentence is annotated by one or several expert models, which we abstract as any model that can show additional data perspectives. For example, these experts could be a terminology list or a statistical phrase-based system to generate translations for specific phrases, but it can also be used in various other situations. For example, we might use it to integrate a model that can do metric conversion or handling of links to web addresses, which can be useful for certain applications. Then NMT model then learns to translate to the target sentence using the annotated source.

\subsection{Annotation}
The aforementioned idea of Experts in our work is inspired by the fact that human translators can benefit from domain experts when translating domain-specific content. Accordingly, we design the annotation and training process as follows: 
\begin{itemize}
\item Words are identified as candidates for annotation using a frequency threshold. 
\item Look up possible translations of the candidates from the Expert and annotate them directly next to the candidates. We use special bounding symbols to help guide the model to copy the annotation during translation. 
\item Train a neural machine translation model using these annotated sentences.
\item During inference, we annotate the source sentence in the same fashion as in training.
\end{itemize}

\paragraph*{Byte-Pair encoding} We consider BPE~\cite{Sennrich2016} one of the crucial factors for annotation in order to efficiently represent words that do not appear in the training data. The rare words (and their translation suggestions, which can be rare as well) are split into smaller segments, alleviating the problem of dealing with $UNK$ tokens~\cite{luong2014addressing}. 

\paragraph*{Embedding sharing} Our annotation method includes target language tokens directly in the source sentence. In order to make the model perceive these words the same way in the source and the target, we create a joint vocabulary of the source and target language and simply tie the embedding projection matrices of the source encoder, target encoder and target decoder. This practice has been explored in various language modeling works~\cite{press2016using,inan2016tying} to improve regularisation.

\subsection{Copy-Generator}
Hypothetically, the model could learn to simply ignore the annotation during optimization because it contains strange symbols (the target language) in source language sentences. If this were the case, adding annotations would not help translate rare events.

Therefore, inspired by the CopyNet~\cite{gu2016incorporating,gulcehre2016pointing}, which originates from pointer networks~\cite{vinyals2015pointer} that learn to pick the tokens that appeared in the memory of the models, we incorporate the copy-mechanism into the neural translation model so that the annotations can be simply pasted into the translation. Explicitly, the conditional probability is now presented as a mixture of two distributions: copy and generated.

\begin{equation}
\label{eq:copy}
\begin{aligned}
P(Y|X; \theta) = \\ \Sigma_{t=1}^T [\gamma P_G(y_t | X, y_1, y_2, \cdots, y_t-1) \\
							+ (1 - \gamma) P_C(y_t |X, y_1, y_2, \cdots, y_t-1)]
\end{aligned}
\end{equation}

The distribution over the whole vocabulary $P_G$ is estimated from the softmax layer using equation~\ref{eq:nmt}, and the copy distribution $P_C$ is used from the attention layer from the decoder state over the context (dubbed `alignment' in previous works~\cite{Bahdanau2014}). The mixture coefficient $\gamma$ controls the bias between the mixtures and is estimated using a feed-forward neural network layer with a sigmoid function, which is placed on top of the decoder hidden state (before the final output softmax layer~\footnote{Using an additional attention layer yields similar result.}). Ideally, the model learns to adjust between copying the input annotation or generating a translation.

It is important to note that, in previous works the authors had to build dynamic vocabulary for each sample due to the vocabulary mismatch between the source and target~\cite{gu2016incorporating}. Since we tied the embeddings of source and target languages, it becomes trivial to combine the two distributions. The use of byte-pair encodings also helps to eliminate unknown words on both sides, alleviating the task of excluding copying unknown tokens.

\subsection{Reinforcement Learning}

\paragraph*{Why reinforcement learning}
While our annotation provides target language tokens that can be directly copied to the generated output, and the copy generator allows a direct gradient path from the output to the annotation, the annotation is not guaranteed to be in the reference. When this is the case, the model does not receive the learning signal to copy the annotation. 

In order to remedy this, we propose to cast the problem as a reinforcement learning task~\cite{ranzato2015sequence} in which we have the model sample and provide a learning signal by rewarding the model if it copies the annotation into the target, as seen in the loss function in Equation~\ref{eq:rl}:

\begin{equation}
\label{eq:rl}
L(\theta) = - \mathbb{E}_{W \sim p_{\theta}}(r(W, REF))
\end{equation}. 
\paragraph*{Reward function} 
For this purpose, we designed a reward function that can encourage the model to prioritize copying the annotation into the target, but still maintain a reasonable translation quality. For suggestion utilization, we denote $HIT$ as the score function that gives rewards for every overlap of the output and the suggestion. If all annotated words are used then $HIT(W, REF)=1.0$, otherwise the percentage of the copied words. For the translation score, we use the GLEU function~\cite{wu2016google} - the minimum of recall and precision of the $n$-grams up to $4$-gram between the sample and the reference, which has been reported to correspond well with corpus-level translation metrics such as BLEU~\cite{papineni2002bleu}. The reward function is defined as in Equation~\ref{eq:reward}:
\begin{equation}
\begin{aligned}
\label{eq:reward}
r(W, REF) = \alpha HIT(W, REF) + \\
			  ( 1  - \alpha ) GLEU(W, REF)
\end{aligned}
\end{equation}


\paragraph*{Variance reduction}
The use of reinforcement learning with translation models has been explored in various works~\cite{ranzato2015sequence,bahdanau2016actor,rennie2016self,nguyen2017reinforcement}, in which the models are difficult to train due to the high variance of the gradients~\cite{schulman2017proximal}. To tackle this problem, we follow the Self-Critical model proposed by~\cite{rennie2016self} for variance reduction:

\begin{itemize}
\item Pre-training the model using cross-entropy loss (Eq.~\ref{eq:nmt}) to obtain a solid initialization pre-search, which allows the model to achieve reasonable rewards to learn faster. 
\item During the reinforcement phase, for each sample/mini-batch, the decoder explores the search space with Markov chain Monte Carlo sampling, and at the same time performs a greedy search for a `baseline' performance. We encourage the model to perform better than baseline, which is used to decide the sign of the gradients~\cite{williams1992simple}.
\end{itemize}

Notably, there is no gradient flowing in the baseline subgraph since the argmax operators used in the greedy search are not differentiable.

\section{Experiment setup}
In the experiments, we realise the generic framework described in Section~\ref{sect:framework} with the tasks of translating from English$\rightarrow$Spanish and German$\rightarrow$English. 

For both language pairs, we used data from Europarl (version 7)~\cite{koehn2005europarl} and IWSLT17~\cite{Cettolo2012WIT} to train our neural networks. For validation, we use the IWSLT validation set (dev2010) to select the best models based on perplexity (for cross-entropy loss) and BLEU score (for reinforcement learning). For evaluation, we use IWSLT tst2010 as the in-domain test set. We also evaluate our models on out-of-domain corpora. For English$\rightarrow$Spanish an additional Business dataset is used.  The corpus statistics can be seen on Table~\ref{tab:stats}. The out-of-domain experiments for the German$\rightarrow$English are carried out on the medical domain, in which we use the UFAL Medical Corpus v1.0 corpus (2.2 million sentences) to train the Expert and the Oracle system. The test data for this task is the HIML2017 dataset with 1517 sentences. We preprocess all the data using standard tokenization, true-casing and BPE splitting with 40K joined operations.

\subsection{Implementation details}
Our base neural machine translation follows the neural machine translation with global attention model described in~\cite{luong2015effective}~\footnote{The framework is implemented in PyTorch and can be found at \textit{https://github.com/quanpn90/OpenNMT-py}}. The encoder is a bidirectional LSTM network, while the decoder is an LSTM with attention, which is a 2-layer feed-forward neural network~\cite{Bahdanau2014}. We also use the input-feeding method~\cite{luong2015effective} and context-gate~\cite{tu2016modeling} to improve model coverage. All networks in our experiments have layer size (embedding and hidden) of $512$ (English$\rightarrow$Spanish) and $1024$ (German$\rightarrow$English) with $2$ LSTM layers. Dropout is put vertically between LSTM layers to improve regularization~\cite{pham2014dropout}. We create mini-batches with maximum $128$ sentence pairs of the same source size. For cross-entropy training, the parameters are optimized using Adam~\cite{kingma2014adam} with a learning rate annealing schedule suggested in~\cite{denkowski2017stronger}, starting from $0.001$ until $0.00025$. After reaching convergence on the training data, we fine-tune the models on the IWSLT training set with learning rate of $0.0002$. Finally, we use our best models on the validation data as the initialization for reinforcement learning using a learning rate of $0.0001$, which is done on the IWSLT set for $50$ epochs. Beam search is used for decoding.

\subsection{Phrase-based Experts}
We selected phrase tables for the Experts in our experiments. 
While other resources like terminology lists can also be used for the translation annotations, our motivation here is that the phrase-tables can additionally capture multi-word phrase pairs, and additionally can better capture the distribution tail of rare phrases as compared to neural models~\cite{koehn2017six}. We selected the translation with the highest average probabilities in the $4$ phrase table scores for annotation.

On the English$\rightarrow$Spanish task, the phrase tables are trained on the same data as the NMT model, while on the German$\rightarrow$English direction, we simulate the situation when the expert is not in the same domain as the test data to observe the potentials. Therefore, we train an additional table on the UFAL Medical Corpus v.1.0 corpus (which is not observed by the NMT model) to for the out-of-domain annotation.

\begin{table*}[!htb]
 \begin{tabular}{lcc|cc}
 & \multicolumn{2}{c|}{\textbf{English$\rightarrow$Spanish}} & \multicolumn{2}{c}{\textbf{German$\rightarrow$English}} \\
 Portion & N. Sentences & Rare words coverage &  N. Sentences & Rare words coverage  \\
 \hline
 All & 2.2M & 82\% (68K) &  1.9M & 82\% (68K) \\
 IWSLT Dev2010 & 1435 & 48\% (135) & 505 & 51\% (196)  \\
 IWSLT Test2010 & 1701 & 46\% (124) & 1460 & 50\% (136) \\
 Out-of-domain & 749 & 80\% (384) & 1511 & 66.64\% (1334) \\
\end{tabular}
\caption{\label{tab:stats} Phrase-table coverage statistics. The out-of-domain section in English-Spanish is Business and Biomedical in German-English. We show the total number of rare words detected by frequency (in parentheses) and the percentage covered by the Experts (intersecting with the reference).}
\end{table*}

\section{Evaluation}

\subsection{Research questions}
\label{ssect:questions}
We aim to find the answers to the following research questions:
\begin{itemize}
\item Given the annotation quality being imperfect, how much does it affect the overall translation quality? 
\item How much does annotation participate in translating rare words, and how consistently can the model learn to copy the annotation?
\item How will the model perform in a new domain? The copy mechanism does not depend on the domain of the training or adaptation data, which is optimal.
\end{itemize}

\subsection{Evaluation Metrics}
To serve the research questions above, we use the following evaluation metrics:
\begin{itemize}
\item BLEU: score for general translation quality. 
\item SUGGESTION (SUG): The overlap between the hypothesis and the phrase-table (on word level), showing how much the expert content is used by the model.  
\item SUGGESTION ACCURACY (SAC): The intersection between the hypothesis, the phrase-table suggestions and the reference. This metrics shows us the accuracy of the system on the rare-words which are suggested by the phrase-table.
\end{itemize}

\paragraph*{Discussion}
The SUG metric shows the consistency of the model on the copy mechanism. Models with lower SUG are not necessarily worse, and models with high SUG can potentially have very low recall on rare-word translation by systematically copying bad suggestions and failing to translate rare-words where the annotator is incorrect. However, we argue that a high SUG system can be used reliably with a high quality expert. For example, in censorship management or name translation which is strictly sensitive, this quality can help reducing output inconsistency. On the other hand, the SAC metrics show improvement on rare-word translation, but only on the intersection of the phrase table and the reference. This subset is our main focus. General rare-word translation quality requires additional effort to find the reference aligned to the rare words in the source sentences, which we consider for future work. 


\subsection{Experimental results}
 \begin{table*}[!htb]
  \begin{center}
   \begin{tabular}{|l|c|c|c|c|c|c|c|c|c|} \hline 
 \hline 
 System|Data & \multicolumn{3}{c|}{\textbf{dev2010}} & \multicolumn{3}{c|}{\textbf{tst2010}} & \multicolumn{3}{c|}{\textbf{BusinessTest}} \\
 & \textbf{BLEU} & \textbf{SAC} & \textbf{SUG} & \textbf{BLEU} & \textbf{SAC} & \textbf{SUG} & \textbf{BLEU} & \textbf{SAC} & \textbf{SUG} \\
  \hline \hline
 \hline
 1. Baseline         & 37.0 & 78.8 & 48.1 & 31.1 & 73.7 & 46.0 & 32.1 & 69.6 & 58.1 \\
  \hline
 2. + AN         	& 37.0 & 97.0 & 71.9 & 31.1 & 93.0 & 74.2 & 32.0 & 91.5 & 79.1 \\
 3. + AN-RF          & \textbf{37.97} & 92.42 & 82.2 & 31.3 & \textbf{94.73} & \textbf{89.5} & \textbf{33.82} & \textbf{96.1} & \textbf{93.0} \\ 
 4. + AN-CP         & 37.3 & 90.9 & 77.8 & 30.7 & 96.5 & 85.5 & 33.2 & 89.8 & 84.9 \\
 5. + AN-CP-RF      & \textbf{38.1} & \textbf{100} & 99.2 & 31.13 & \textbf{100} & \textbf{99.2} & 33.34 & \textbf{98.3} & \textbf{97.6} \\ \hline
   \end{tabular}
 \caption{\label{result:enes} The results of English - Spanish on various domains: TEDTalks and Business. We use AN for using annotations from the phrase table, RF for using REINFORCE ($\protect\alpha$= 0.5) and CP for using the Copy mechanism.}
 
  \end{center}
\end{table*}

 \begin{table*}[!htb]
  \begin{center}
   \begin{tabular}{|l|c|c|c|c|c|c|c|c|c|} \hline 
 \hline 
 System|Data & \multicolumn{3}{c|}{\textbf{dev2010}} & \multicolumn{3}{c|}{\textbf{tst2010}} & \multicolumn{3}{c|}{\textbf{HIML}} \\
 & \textbf{BLEU} & \textbf{SAC} & \textbf{SUG} & \textbf{BLEU} & \textbf{SAC} & \textbf{SUG} & \textbf{BLEU} & \textbf{SAC} & \textbf{SUG} \\
  \hline \hline
 \hline
 1. Baseline         & \textbf{37.5} & 66 & 45 & \textbf{36.14} & 66.9 & 45.1 & 32.4 & 46.3 & 37.2 \\
  \hline
 2. + AN         	& 37.1 & 93 & 84.1 & 35.6 & 91.9 & 84.4 & 33.99 & 87.1 & 85.1 \\
 3. + AN-CP         & 37.2 & 96 & 88.2 & 35.89 & 94.1 & 90.7 & 34.1 & \textbf{96.5} & \textbf{95.0} \\
 4. + AN-CP-RF      & 36.6 & \textbf{97} & \textbf{92.9} & 35.89 & \textbf{98.5} & \textbf{95.5} & 33.1 & \textbf{98.0} & \textbf{97.6} \\ \hline
 Biomedical-Oracle & - & - & - & - & - & - & \textbf{37.82} & 81.77 & 65.44 \\ \hline
   \end{tabular}
 \caption{\label{result:deen} The results of German$\rightarrow$English on various domains: TEDTalks and Biomedical. We use AN for using annotations from the phrase table, RF for using REINFORCE ($\protect\alpha$= 0.5) and CP for using the Copy mechanism.}
 
  \end{center}
\end{table*}

\paragraph*{English$\rightarrow$Spanish} Results for this task are presented on table~\ref{result:enes}. First, the main difference between the settings is the SUG and SAC figures for all test sets. Both of them increase dramatically from baseline to annotation, and also increase according to the level of supervision in our model proposals. While the copy mechanism can help us to copy more from the annotation, the REINFORCE models are successfully trained to make the model copy more consistently. Their combination helps us achieve the desired behavior, in which almost all of the annotations given are copied, and we achieve 100\% accuracy on the rare-words section that the phrase table covers. As mentioned in the discussion above, the SAC and SUG figures, while being not enough to quantitatively prove that the total number of rare words translated, show that the phrase table is complementary to the neural machine translation, and the more coverage the expert has, the more benefit this method can bring. 

We notice an improvement of 1 BLEU point on dev2010 but only slight changes compared to the baseline on tst2010. On the out-of-domain set, however, the improved rare-word performance leads to an increase of $1.7$ BLEU points over the baseline without annotation. Our models, despite training on a noisier dataset, are able to improve translation quality. 

\paragraph*{German$\rightarrow$English} Results are shown in Table~\ref{result:deen}. On the dev2010 and tst2010 in-domain datasets, we observe similar phenomena to the En-Es direction. Rare-word performance increases with the number of words copied, and the combination of the copy mechanism and REINFORCE help us copy consistently. Surprisingly, however, the BLEU score drops with annotations. This may be because of the relative morphologically complexity of the German words compared to the English, making it harder to generate the correct word form.

In the experiments with an out-of-domain test set (HIML), we use annotations from that domain to simulate a domain-expert. For comparison, we also trained an NMT model adapted to the UFAL corpus, which we call the Oracle model. In this domain, our models show the same behavior, in which almost every word annotated is copied to the output. The annotation efficiently improves translation quality by 1.7 BLEU points over the baseline without annotation. The adapted model has a higher BLEU score, but here performs worse than our annotated model in terms of phrase-table overlap and rare-word translation accuracy for words in this set. Our model shows significantly better rare word handling than the baseline. Though the best obtainable system is adapted to the in-domain data, this requires parallel text: this experiment shows the high potential to improve NMT on out-of-domain scenarios using only lexical-level materials. We notice a surprising drop of 1.0 BLEU points for the REINFORCE model. Possible reasons include inefficient beam search on REINFORCE models, or the GLEU signal was out-weighted by the HIT one during training, which is known for the difficulty~\cite{zaremba2015reinforcement}.

\subsection{Further Analysis}
\begin{figure*}[htb]
\caption{\label{fig:example1} \textbf{Top}: Examples of name annotations with our framework from tst2010. The name Kean is originally split by BPE into `K' and `ean'. This is incorrectly translated without annotation (in blue) and corrected with the annotation (in red). \textbf{Bottom}: An example of phrase copying, in which the German word is translated into a long English phrase.}
\centering
\includegraphics[scale=0.25]{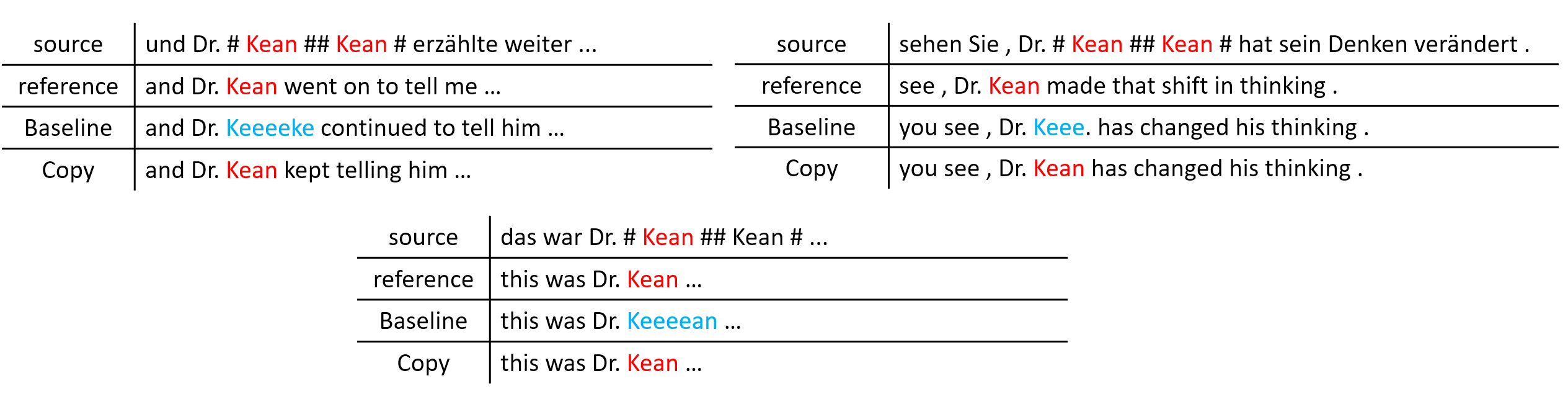}
\vline
\includegraphics[scale=0.25]{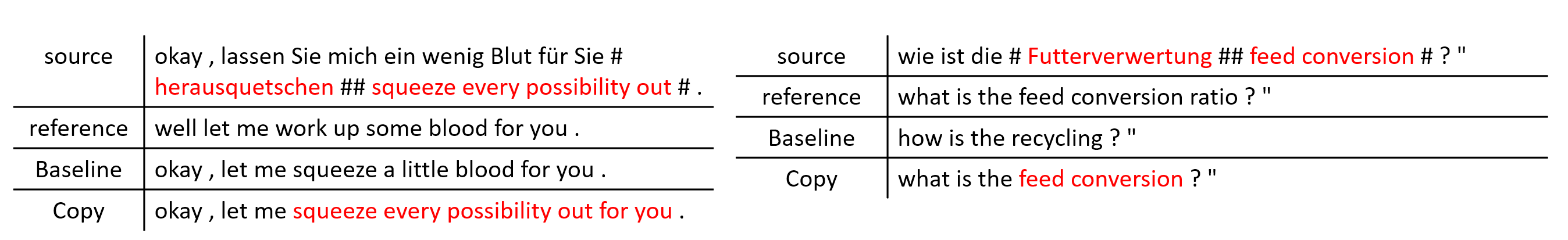}

\end{figure*}

\paragraph{Name translation} Names can often be translated by BPE, but it is noticeable about  examples of the inconsistency, which can be alleviated using annotations, as illustrated in Figure~\ref{fig:example1}-Top.

\paragraph{Copying long phrases} We find that with very high supervision, the model can learn to copy even phrases completely into the output, as in Figure~\ref{fig:example1}-Bottom. Though this is potentially dangerous, as the output may the lose the additional fluency which comes from NMT, it is controllable by combining RL and cross entropy loss~\cite{paulus2017deep}. 

\begin{figure*}[htb]
\vspace{-1em}
\caption{\label{fig:attn} An attention heat map of an English-Spanish sentence pair (source on X-axis, target on Y-axis) with annotated sections in red rectangles. Annotations and their source are bounded by \# characters.}
\centering
\includegraphics[width = 0.65\textwidth]{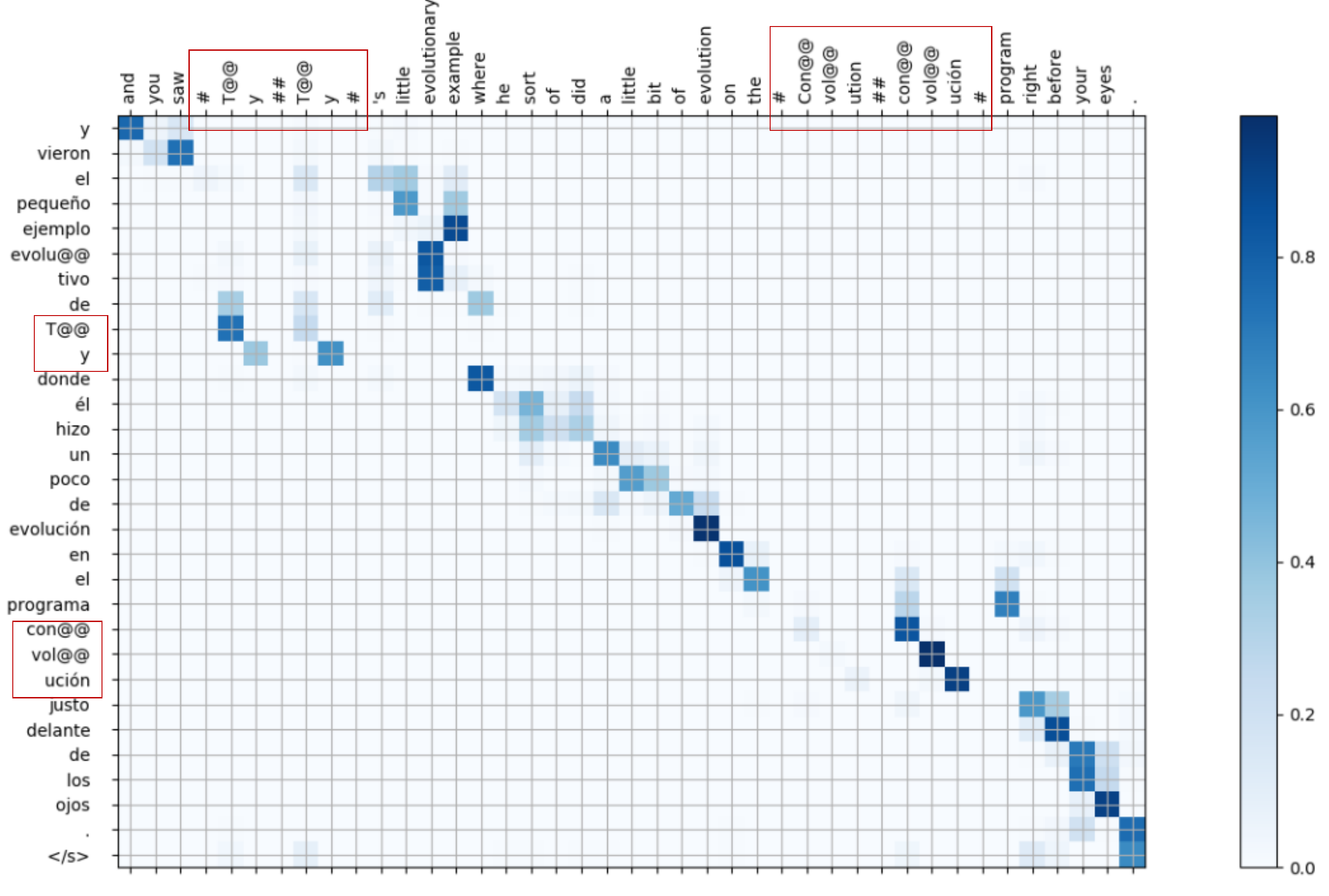}
\end{figure*}

\paragraph{Attention} Plotting the attention map for the decoded sequences we notice that, while we marked the beginning and end of annotated sections and the separation between the source and the suggestion with \# and \#\# tokens, those positions received very little weight from the decoder. 
One possible explanation is that these tokens do not contribute to the translation when decoding, and the annotations may useful without bounding tags. For the annotations used in the translation, we identified two prominent cases; for the rare words whose annotation need only be identically copied to the target, the attention map focuses evenly on both source and annotation, while the heat map typically heavily emphasizes only the annotation otherwise. An example is illustrated in figure~\ref{fig:attn}.

\paragraph{Effect of $\alpha$}
The full results with respect to different $\alpha$ values which are used in Equation~\ref{eq:rl} for reward weighting can be seen in Table~\ref{tab:alpha}. Higher $\alpha$ values emphasize the signal to copy the source annotation, as can be seen from the increase in terms of Accuracy and Suggestion utilization across the values. As expected, as $\alpha$ goes toward 1.0, the model gradually loses the signal needed to maintain translation quality and finally diverges. 

 \begin{table}[!htb]
  \begin{center}
   \begin{tabular}{|l|ccc|ccc|} 
 \hline 
 $\alpha$ & \multicolumn{3}{c|}{\textbf{tst2010}} & \multicolumn{3}{c|}{\textbf{BusinessTest}} \\
 & \textbf{BL} & \textbf{AC} & \textbf{SUG} & \textbf{BL} & \textbf{AC} & \textbf{SUG} \\
 0.0 & 31.3 & 91.2 & 78.2 & 33.7 & 76.5 & 71.9 \\
 0.2 & 31.0 & 94.7 & 88.2 & 33.9 & 78.1 & 76.8 \\
 0.5 & 31.3 & 94.7 & 89.5 & 33.8 & 96.1 & 93.0 \\
 1.0 & \multicolumn{6}{c|}{did not converge} \\	
 \hline
   \end{tabular}
 \caption{\label{tab:alpha} Performances w.r.t to different alpha values. Metrics shown are BLEU (BL), ACCURACY (AC) and SUGGESTION (SUG)}
  \end{center}
 \end{table}
 \vspace{-1.5em}

\section{Related Work}
\paragraph{Translating rare words} in neural machine translation is a rich and active topic, particularly when translating morphologically rich languages or translating named entities. Sub-word unit decomposition or BPE ~\cite{Sennrich2016} has become the de-facto standard in most neural translation systems~\cite{wu2016google}. 
Using phrase tables to handle rare words was previously explored in~\cite{luong2014addressing}, but was not compatible with BPE. 
\cite{gulcehre2016pointing}~explored using pointer networks to copy source words to the translation output, which could benefit from our design but would require significant changes to the architecture and likely be limited to copying only. Additionally, models that can learn to remember rare events were explored in~\cite{kaiser2017rare}. 

Our work builds on the idea of using a phrase-based neural machine translation to augment source data,~\cite{niehues2016pre,denkowski2017stronger}, but can be extended to any annotation type without complicated hybrid phrase-based neural machine translation systems. We were additionally inspired by the use of feature functions with lexical-based features from dictionaries and phrase-tables in~\cite{zhang2017prior}. They also rely on sample-based techniques,~\cite{shen2015minimum}, to train their networks, but their computation is more expensive than the self-critical network in our work. We focus here on rare events, with the possibility to construct interactive models for fast updating without retraining.
We also use the ideas of using REINFORCE to train sequence generators for arbitrary rewards~\cite{ranzato2015sequence,nguyen2017reinforcement,bahdanau2016actor}. While this method remains difficult to train, it is promising to use to achieve non-probabilistic features for neural models: for example enforcing formality in outputs in German, or censoring undesired outputs.

\section{Conclusion}
In this work, we presented a framework to alleviate the weaknesses of neural machine translation models by incorporating external knowledge as~\textbf{Experts} and training the models to use their annotations using reinforcement learning and a pointer network. We show improvements over the unannotated model on both in- and out-of-domain datasets. When only lexical resources are available and in-domain fine-tuning cannot be performed, our framework can improve performance. The annotator might potentially be trained together with the main model to balance translation quality with copying annotations, which our current framework seems to be biased to.
\vspace{-0em}

\section*{Acknowledgments}
\vspace{-0.5em}
This work was supported by the Carl-Zeiss-Stiftung. We thank Elizabeth Salesky for the constructive comments.

\bibliography{references.bib}
\bibliographystyle{acl_natbib}

\appendix

\end{document}